\documentclass[11pt,a4paper]{article}
\usepackage[hyperref]{emnlp-ijcnlp-2019}
\usepackage{times}
\usepackage{latexsym}
\usepackage{multirow}
\usepackage{amsfonts}
\usepackage{CJKutf8}
\usepackage{epsfig}
\usepackage{color}
\usepackage{graphicx}
\usepackage{amsfonts}
\usepackage{amsmath}
\usepackage{bm}
\usepackage{algorithmicx}
\usepackage{url}
\usepackage{footnote}
\usepackage{bbm}
\usepackage{url}
\usepackage[linesnumbered,ruled,commentsnumbered]{algorithm2e}

\aclfinalcopy

\title{Multi-agent Learning for Neural Machine Translation}

\author{
 Tianchi Bi, Hao Xiong, Zhongjun He, 
  Hua Wu and \textbf{Haifeng Wang} \\
 Baidu Inc. No. 10, Shangdi 10th Street \\
 Beijing, 100085, China \\
  {\sf \{bitianchi, xionghao05,  hezhongjun, wu\_hua, wanghaifeng\}}@baidu.com}

\date{}

\begin{document}
\maketitle
\begin{CJK}{UTF8}{gbsn}
\begin{abstract}
Conventional Neural Machine Translation (NMT) models benefit from the training with an additional agent, e.g., dual learning, and bidirectional decoding with one agent decoding from left to right and the other decoding in the opposite direction. In this paper, we extend the training framework to the multi-agent scenario by introducing diverse agents in an interactive updating process. At training time, each agent learns advanced knowledge from others, and they work together to improve translation quality. Experimental results on NIST Chinese-English, IWSLT 2014 German-English, WMT 2014 English-German and large-scale Chinese-English translation tasks indicate that our approach achieves absolute improvements over the strong baseline systems and shows competitive performance on all tasks.
\end{abstract}

\section{Introduction}
Training with more than one agents has attracted intensive research interest in recent years, for example, dual learning \cite{he2016dual,xia2017dual,xia2018model} and bidirectional decoding \cite{liu2016agreement,zhang2018regularizing}.  The former method leverages
the duality between the two related agents as the feedback signal to regularize training, while the latter targets the agreement between one agent decoding from left to right (L2R) while the other decoding in opposite direction (R2L). Both  methods enhance the translation models by introducing a regularization term into the training objective. 

The effectiveness of these two methods lies in the fact that appropriate regularization can help each agent to learn from the superior models while integrating their advantages (e.g., good translations for prefixes for L2R, and good translation quality for suffixes for R2L). 
As shown in the Table \ref{tbl:inst}, due to the \textit{exposure bias} problem \cite{ranzato2015sequence}, the agent trained to decode from left to right tends to generate better prefixes and bad suffixes, and the agent decoding in the reverse direction demonstrates the opposite preference. By introducing additional Kullback-Leibler (KL) divergences between the probability distributions defined by L2R and R2L models into the NMT training objective \cite{zhang2018regularizing}, it is possible for two models to learn advantages from each other.

\begin{table}[t]

\begin{center}
\begin{tabular}{p{0.6 cm}|p{0.82\columnwidth}}

\textit{Src} & 此次 豪雨 灾情 , 半数 遇难 者 遭 洪水 冲走 溺 毙 ; 其他 则 因 房屋 倒塌 或 电线 走火 触电 致死 。 \\

\hline
\textit{Ref} & In this disaster caused by torrential rains, half of the victims were carried away and drowned in the floods; the others died because their houses collapsed or from electrocution caused by short circuits. \\

 \hline
 \textit{L2R} & In the torrential rain, half of the victims were washed away by floods, while others \textcolor{red}{died of electricity caused by collapsed houses or electric wires.} \\
 \hline
 \textit{R2L} & \textcolor{red}{Half of the victims were drowned by floods,} while others were killed by collapsed houses or electrocuted by sparkling wires. \\
  
\end{tabular}
\end{center}
\caption{In this sample, NMT systems with different implementations present diverse translation errors for one long sentence. The agent decoding from left to right (\textit{L2R}) tends to generate better prefixes and bad suffixes (red letters), and the \textit{R2L} agent has the opposite preference. }
\label{tbl:inst}
\end{table}

According to the empirical achievements of previous studies on training with two agents, it is natural to consider the training with more than two agents, and to extend our study to the multi-agent scenario. However, training with more than two agents is more complex, and we face two critical problems. First, when deploying the multi-agent system, should we focus on the diversity or the strength of each agent? \cite{wong2005effects,marcolino2013multi} Second, learning in multi-agent scenario is \textit{many-to-many}, as opposed to the relatively simpler \textit{one-to-one} learning in two-agent training, and requires an effective learning strategy.  

There have been many alternatives to improve the diversity of models even based on the Transformer model \cite{vaswani2017attention}.
For example, decoding in the opposite direction usually results in different preferences: good prefixes and bad prefixes \cite{zhang2018regularizing}. Rather, self-attention with relative position representations enhances the generalization to sequence lengths unseen during training \cite{shaw2018self}. Furthermore, increasing the size of layers in the encoder is expected to specialize in word sense disambiguation \cite{D18-1458,domhan2018much}. In this paper, we investigate the effects of two teams of multi-agents: a team of alternative agents mentioned above, and a uniform team of different initialization for the same model. 
 
To resolve the second problem, we simplify the \textit{many-to-many} learning to the \textit{one-to-many} (one teacher $vs.$ many students) learning, extending ensemble knowledge distillation \cite{fukuda2017efficient, freitag2017ensemble,P18-1129, zhu2018knowledge}. During the training, each agent performs better by learning from the ensemble model (\textit{Teacher}) of all agents integrating the knowledge distillation \cite{44873,kim2016sequence} into the training objective. This procedure can be viewed as the introduction of an additional regularization term into the training objective, with which each agent can learn advantages from the ensemble model gradually. With this method, each agent is optimized not only to the maximization of the likelihood of the training data, but also to the minimization of the divergence between its own model and the ensemble model.

However, the above learning strategy converges on the local optimum rapidly in our empirical studies. 
It seems that each agent tends to focus on learning from the ensemble model while completely ignoring its own exploration.  
To alleviate this problem, we train each agent to learn from the ensemble models when necessary, and to distill the knowledge based on the translation quality (BLEU score) of the ensemble model. This means we evaluate the quality of the ensemble model to see whether it is good enough to be studied.
Consequently, the knowledge distillation from the ensemble model to the agent stems from the better translation generated by the ensemble model.

%Otherwise, the agent is forced to learn its own distribution as in REINFORCE learning \cite{williams1992simple}. 
%This learning strategy can be related to the study on learning to teach \cite{fan2018learning}, that improves the learning %efficiency in the \textit{Teacher}-\textit{Student} learning scenario.  

%Lastly, we integrate optimization of each model into a joint training framework, in which each agent acts as helper system for other agents, and all agents achieve further improvements with an interactive updating process.

We evaluate our model on NIST Chinese-English Translation Task, IWSLT 2014 German-English Translation task, large-scale WMT 2014 English-German Translation task and large-scale Chinese-English Translation Task. Extensive experimental results indicate that our model significantly improves the translation quality, compared to the strong baseline systems. Moreover, our model also reports competitive performance on the German-English and English-German translation tasks, achieving 36.27 and 29.67 BLEU scores, respectively.

To the best of our knowledge, this is the first work on training NMT model with more than two agents 
\footnote{Although \citet{dual2019} proposed a work named multi-agent dual learning contemporarily, the training in their work is conducted on the two agents with fixed parameters of other agents.}. 
The contributions of this paper are summarized as follows:
\begin{itemize}
\item We extend the study on training with two agents to the multi-agent scenario, and propose a general learning strategy to train multiple agents.
\item We investigate the effects of the diversity and the strength of each agent in the multi-agent training scenario.
\item We simplify complex \textit{many-to-many} learning in multi-agent learning to \textit{one-to-many} learning by forcing each agent learning knowledge from ensemble model as necessary.
\item Extensive experiments on multiple translation tasks confirm that our model significantly improves the translation quality and reports the new state-of-the-art results on IWSLT 2014 German-English and competitive results on WMT 2014 English-German translation tasks. 
\end{itemize}

\section{Background}
Conventional autoregressive NMT models \cite{bahdanau2014neural,sutskever2014sequence} decode target tokens sequentially, which indicates that the determination of the current token is conditioned by the previous generated sequence. Formally, at time step $t$, the generation of the current token $y_t$ is determined by the following equation:
\begin{equation}
p(y_t)=p(y_t|y_{<t}, x; \theta)
\label{eqn:py_t}
\end{equation}
where $y_{<t}$ represents the previously generated sequence, and $\theta$ are the parameters of the representation of the source sequence and the partial target sequence. 

Furthermore, the usual training criterion is to minimize the negative log-likelihood for each sample from the training data,
\begin{equation}
\mathcal{L}_{NLL}=-\sum_{t=0}^{T}\sum_{k=1}^{|\mathcal{V}|}{\mathbbm{1}\{y_t=k\}\mathrm{log}p(y_t=k|y_{<t}, x; \theta)}
\label{eqn:loss1}
\end{equation}
where $T$ is the length of the target sequence, ${|\mathcal{V}|}$ is the size of vocabulary and $\mathbbm{1}\{\cdot\}$ is the indicator function. This objective can be viewed as minimizing the cross-entropy between the correct translation $y^*$ and the model generation $p(y_t|y_{<t}, x; \theta)$.

\section{Multi-agent Learning}

\begin{figure}[t!]
\centering
\includegraphics[width=3.0in]{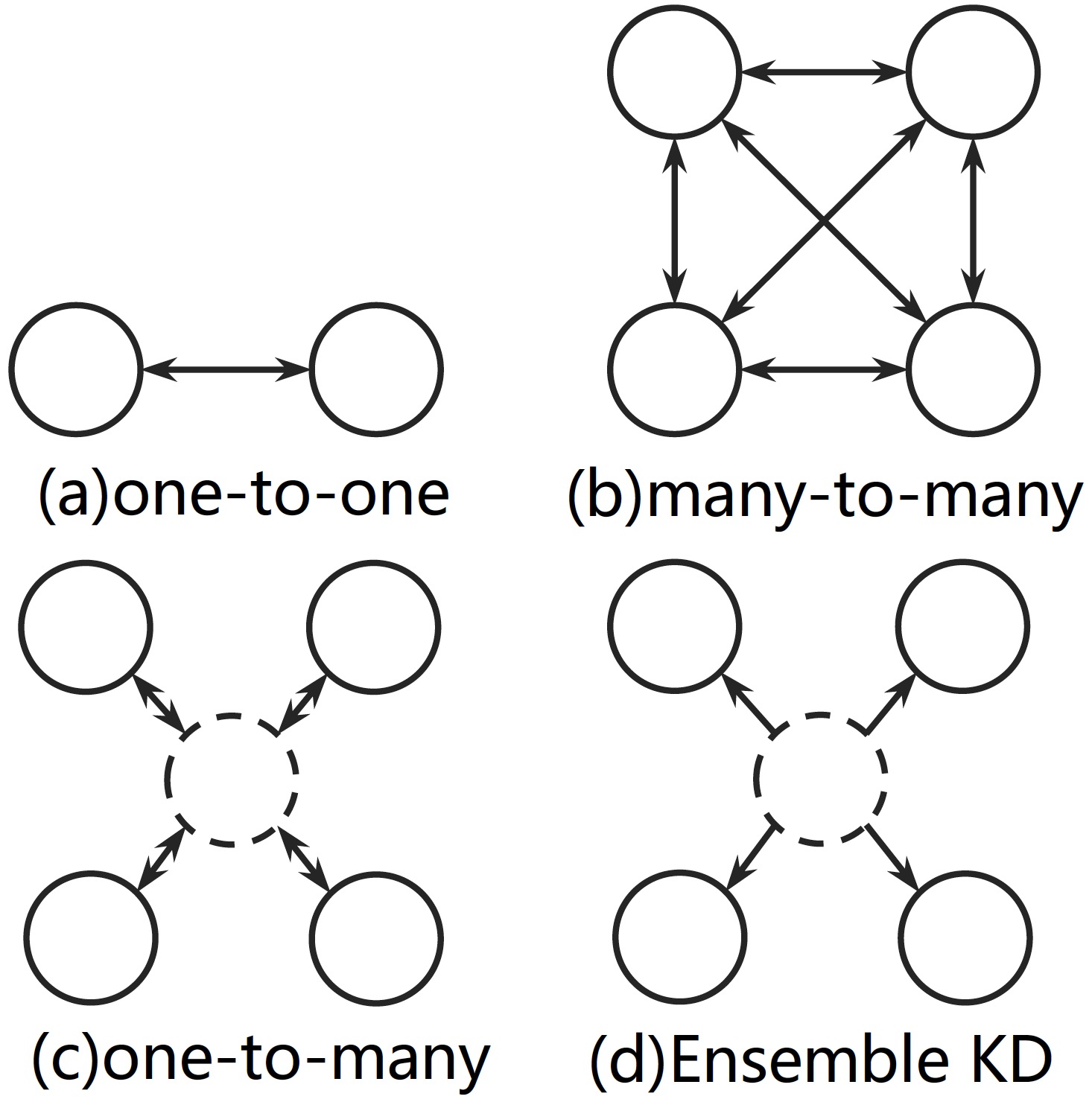}

\caption{Illustration for different learning approaches. }
\label{fig:learning}
\end{figure}

Empirical studies indicate that one agent is trained to perform better through learning advantages from the other agent in the two-agent scenario, namely \textit{one-to-one} learning (Figure \ref{fig:learning}.(a)). The training objective of the agent is regularized, which is to learn better models by leveraging the relationship between the two agents as feedback, i.e., duality in dual learning problem, and agreement in bidirectional decoding.  

Extending the study to the multi-agent scenario is feasible and desirable, as multiple agents might supply more reliable and diverse advantages compared to the two-agents scenario. However, the agent is expected to learn advantages from each other in the multi-agent scenario, which results in a complex \textit{many-to-many} learning problem(Figure \ref{fig:learning}.(b)). 

Instead of tackling \textit{many-to-many} learning, we force the agent to learn from a common \textit{Teacher} by introducing ensemble knowledge distillation, thus reduce the learning to \textit{one-to-many} (Figure \ref{fig:learning}.(c)). With this learning strategy, each agent can learn to improve the performance in an interactive updating process.     

As opposed to ensemble Knowledge Distillation (KD) (Figure \ref{fig:learning}.(d)), the important difference is that in the knowledge distillation of an ensemble of models, the \textit{Teacher} network is fixed after pre-training, during the training process. While in our framework, the state of the \textit{Teacher} network is updated at each iteration, and its performance can be further improved by the improvements of each agent explicitly, in an interactive updating process.  
In some ways the ensemble KD can be viewed as a particular case of our model, as we fix the update of the \textit{Teacher} network in the training framework.

\subsection{Overall Framework}
\begin{figure*}[ht!]
\centering
\includegraphics[width=6.2in]{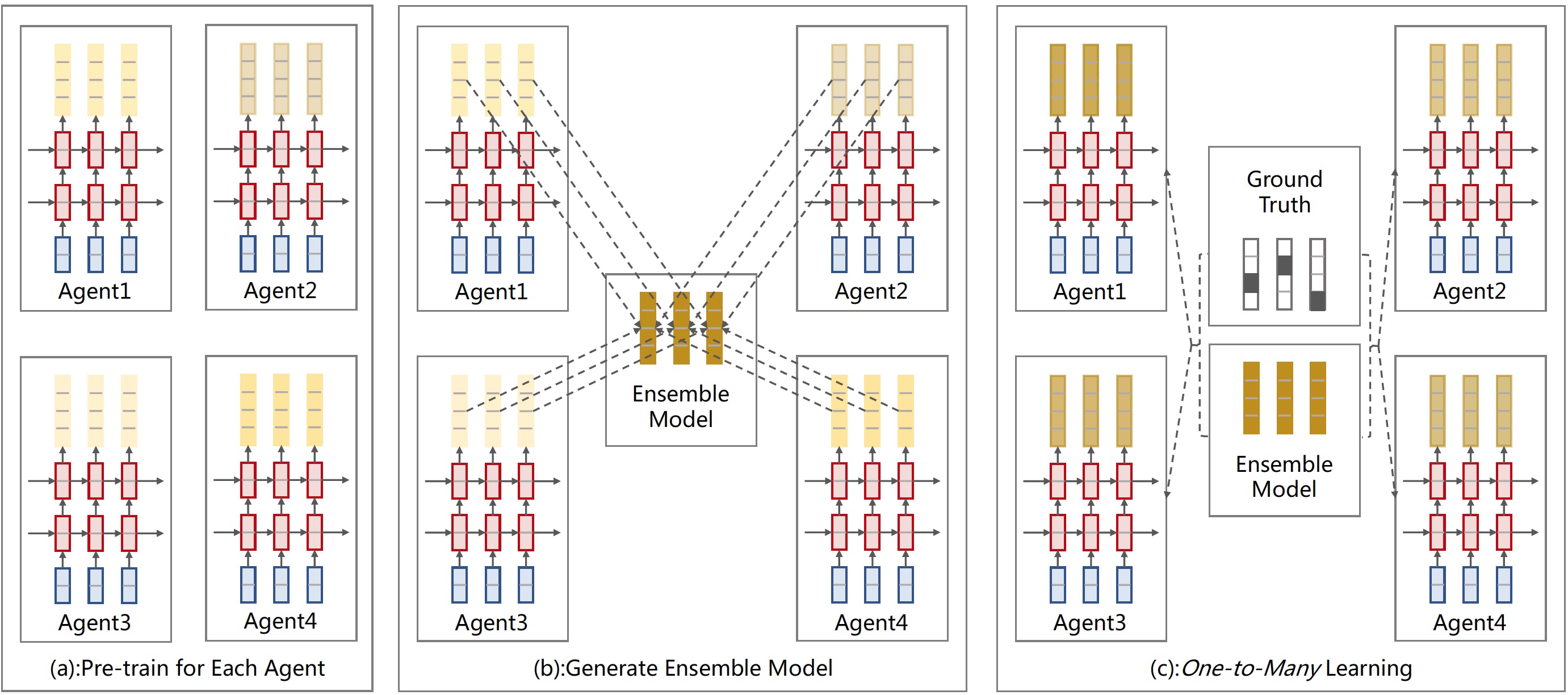}

\caption{In this example, four agents decode the similar sentence with different model capacity. (a): At first, each agent is pre-trained to generate the translation independently. (b) The ensemble model is generated by the average prediction from each agent. (c): The \textit{One-to-Many} learning distills the knowledge from the ensemble model to each agent as necessary. The performance of each agent is improved explicitly in an interactive updating process, through repeating the process (b) and (c). }
\label{fig:arch}
\end{figure*}

For the sake of simplicity, four variant agents are referred to as $Agent1$, $Agent2$, $Agent3$ and $Agent4$. 
We begin by pre-training each agent independently (Figure \ref{fig:arch}.(a)), and then enhance the model in the multi-agent scenario in two steps: 1) Generating Ensemble Model (Figure \ref{fig:arch}.(b)), and 2) \textit{One-to-Many} Learning (Figure \ref{fig:arch}.(c)). The performance of each agent is improved in an interactive updating process, through repeating the above two steps. 

\subsection{Generate Ensemble Model}

As pointed out in the work of \citet{ P18-1129}, ensemble models can empirically alleviate problems existing in the standard NMT model, such as ambiguities in the training data, and discrepancy between training and testing. 

According to the practical advantages of ensemble models, it is relevant to force agents to learn from the ensemble model, instead of learning from each other separately.  
Following previous work, we develop our ensemble model by averaging the model distributions of all agents.

Formally, the model distribution of the $i$-th agent $agent_i$ is defined as $p(y_t^i|y_{<t}^i, x;\theta)$.
Assume we have $\mathcal{N}$ agents, the model distribution of the ensemble model can be formulated as:
\begin{equation}
q(y_t|y_{<t}, x;\theta_t) = \frac{1}{\mathcal{N}}\sum_{i=1}^\mathcal{N}{p(y_t^i|y_{<t}^i, x;\theta_i)}
\label{eqn:ensemble}
\end{equation}
where $\theta_t$ are parameters for representing the ensemble model. Notably, we do not train it in the training process.

In the above formula, the probability $q(y_t|y_{<t}, x;\theta_t)$ is one reliable estimator of the model distribution, as the majority of the agents are likely to generate correct sequence. From this perspective, we expect that more agents will lead to better and more robust performance.
\subsection{\textit{One-to-Many} Learning}
In the \textit{one-to-many} learning framework, the ensemble model acts as the \textit{Teacher} network, which distills knowledge to each agent iteratively.  

Rather than minimizing
cross-entropy with the observed data, we minimize the cross-entropy with the probability distribution from the ensemble model for each agent.
\begin{equation}
\begin{aligned}
\mathcal{L}_{KD}^{i}=&-\sum_{t=0}^{T}\sum_{k=1}^{|\mathcal{V}|}{q(y_t=k|y_{<t}, x;\theta_t)} \times\\
&\mathrm{log}p(y_t^{i}=k|y_{<t}^{i}, x;\theta)
\end{aligned}
\label{eqn:loss1}
\end{equation}
With the above formula, the agent is optimized to the minimization of the model divergence between its own model and the ensemble model. %This regularization term forces the agent to learn advantages from the ensemble model, learning diverse preferences from all agents implicitly as the ensemble model is a reliable estimator of all agents. 

However, integrating the above regularization term into the training objective straightforwardly is problematic in practice. The agent tends to focus on learning from the ensemble model rather than exploring its own prediction to converge rapidly. 
Consequently, the model converges at a suboptimal point,
and fails to enhance the performance by learning from each other due to the lack of evaluation of the ensemble model. 

To alleviate this problem, we train each agent to learn from the ensemble model when necessary, distilling the knowledge conditioned by the translation quality of the ensemble model. We evaluate the quality of the ensemble model to see whether it is good enough to be studied. The knowledge distillation from the ensemble model to the agent comes from the better translation generated by the ensemble model. Otherwise, the agent is forced to learn its own distribution. 

Let $<X_{g}, Y_{g}>$ be one sentence pair in the training corpus, the quality of the translation sequence generated by the ensemble model is measured by the BLEU score,
\begin{equation}
score(Y_{t}) = \mathrm{BLEU}(Y_{t}, Y_{g})
\label{eqn:bleu_teacher}
\end{equation}   
where $Y_{t}$ is generated by the model distribution $q(y_t|y_{<t}, x;\theta_t)$.
The quality of the translation sequence generated by each agent can also be defined using the similar metric,
 \begin{equation}
score(Y_{s}^{i}) = \mathrm{BLEU}(Y_{s}^{i}, Y_{g})
\label{eqn:bleu_agent}
\end{equation} 
where $Y_{s}^{i}$ is generated by the model distribution $p(y_t^i|y_{<t}^i, x;\theta)$ of each agent.

Conditioned by the translation quality of the ensemble model, we modify the training objective for each agent as follows:
\begin{equation}
\mathcal{L}_{KD}^{i}=-\sum_{t=0}^{T}\sum_{k=1}^{|\mathcal{V}|}S \times \mathrm{log}p(y_t^{i}=k|y_{<t}^{i}, x;\theta)
\label{eqn:newkd}
\end{equation}
where $S$ is defined as:
\begin{equation}
S=
\begin{cases}
q(y_t=k|y_{<t}, x;\theta_t)& score(Y_{t}) > score(Y_{s}^{i})\\
\mathbbm{1}_s{\{y_t^{i}=k\}}& otherwise
\end{cases}
\end{equation}
where $\mathbbm{1}_s\{\cdot\}$ is an indicator, conditioned by the existence of $y_t^{i}=k$ in the sequence $Y_{s}^{i}$. 

In practice, each agent is not only optimized to maximize the likelihood of the training data, but also to minimize the model divergence between its own model and the ensemble model:
\begin{equation}
\mathcal{L}_a^i = \lambda_i \mathcal{L}_{NLL} + ( 1 - \lambda_i) \mathcal{L}_{KD}^{i}
\label{eqn:final_loss}
\end{equation}
where $\lambda_i$ is a hyperparameter, balancing the weight of two factors.

From the perspective of learning, the agent learns to minimize the training objective by generating competitive translations, which leads to a global improvement for all agents. On the other hand, a sequence-level training objective might alleviate the \textit{exposure bias} problem implicitly \cite{ranzato2015sequence}. 
\subsection{Joint Learning}
%\IncMargin{1em} % 使得行号不向外突出 
\begin{algorithm}[t]

    %\SetAlgoNoLine % 不要算法中的竖线
    %\SetKwInOut{Input}{\textbf{Input}}\SetKwInOut{Output}{\textbf{Output}} % 替换关键词

    \KwIn{
        $N$ variant agents\;
        }
    
    %\BlankLine

    Pre-train each agent $agent_i$ independently\; % 分号 \; 区分一行结束
    
    \Repeat
        {\text{convergence}}
        {
            \For {each training sample $(X_{g}, Y_{g})$ }{
            \label{lbl:sample}
                
                Ensemble Model: $q(y_t|y_{<t}, x;\theta_t)$\;
                Generate sequence: $Y_t \leftarrow \mathop{\arg\max}\limits_{0 \le t < T} q(y_t|y_{<t}, x;\theta_t)$\;
                \For {each agent $agent_i$}
                {
                	Generate sequence: $Y_s^{i} \leftarrow \mathop{\arg\max}\limits_{0 \le t < T} p(y_t^i|y_{<t}^i, x;\theta)$\;
		
                	Compute KD $loss$: $\mathcal{L}_{KD}^{i} \leftarrow (q(y_t),p(y_t^{i}),Y_t, Y_s^{i}, X_{g},Y_{g})$\; 
		Compute NLL $ loss$: $\mathcal{L}_{NLL} \leftarrow (p(y_t^{i}), X_g, Y_g)$ \;
	 	Compute Agent $loss$: $\mathcal{L}_a^i \leftarrow \lambda \mathcal{L}_{NLL} + ( 1 - \lambda) \mathcal{L}_{KD}^{i}$\;
	  		
                }
                Model $loss$: $\mathcal{L}_{final} = \sum_{i=1}^{\mathcal{N}}{\mathcal{L}_{a}^i}$\;
                Update gradients for each agent\;
            }
           
        }
    \caption{Multi-agent Learning}
    \label{algo1}
\end{algorithm}
In the work of \citet{zhang2018regularizing}, they proposed a relatively complex joint learning framework for training two agents. In this paper, according to the previous experiments, we find that a simple multi-task learning technique without sharing any modules presents promising performance,
\begin{equation}
\mathcal{L}_{final} = \sum_{i=1}^{\mathcal{N}}{\mathcal{L}_{a}^i}
\label{eqn:loss_final}
\end{equation}   

In Algorithm \ref{algo1}, we describe the overall procedure of our approach. It deserves noting that in line \ref{lbl:sample}, we assume the model reads a single pair of training examples per timestep for simplistic description, while in practice the model reads a batch size of samples at each step.

\begin{table*}[thb]
\begin{center}
\begin{tabular}{l|c|c|c|c|l|l}
 MODEL& MT02& MT03& MT04& MT08 & AVERAGE & IMPROVEMENTS \\
 \hline
 & \multicolumn{4}{c|}{Results for Best Agent} & \multicolumn{2}{c}{Results for each Agent} \\
\hline
\textit{\citet{wang2018neural}} &- &46.60 &47.73 &- &- &-\\
 \textit{a.L2R} &48.53 &47.07 &48.43 &42.21 &46.56 &- \\
 \textit{b.R2L} &47.06 &45.58 &47.14 &41.04 &45.20 &-    \\
 \textit{c.Enc} &48.86 &47.54 &48.57 &42.93 &\textbf{46.97} &- \\
 \textit{d.Rel} &48.12 &48.19 &48.33 &42.51 & 46.78 &- \\
 \hline
 \textit{a+b} &48.82  &47.65 &48.45 &42.49 &\textbf{46.86}/45.42 &\textbf{+0.33}/+0.20  \\
 \textit{a+c} &48.79  &48.30  &49.32 &43.44 &\textbf{47.3}/47.27 &\textbf{+0.80}/+0.30 \\
 \textit{a+d} &48.76 &48.40 &48.74 &43.27 &\textbf{47.19}/47.09  & \textbf{+0.63}/+0.31 \\
 \textit{c+d} &49.45 &49.01 &49.52 &43.71 &47.62/\textbf{47.79} &+0.65/\textbf{+1.01}\\
 \hline
 \textit{a$\times$2} &48.64  &47.98 &49.08 &43.07 &46.96/\textbf{47.10} &+0.40/\textbf{+0.54}  \\
  \textit{d$\times$2} &48.23  &48.78 &48.90 &43.85 &\textbf{47.44}/47.31 &\textbf{+0.66}/+0.53  \\
 \hline
  \textit{a+b+c} &49.32  &48.72 &49.32 &44.34 &47.69/45.62/\textbf{47.74} & \textbf{+1.13}/+0.42/+0.77\\
 \textit{a+b+d} &49.29 &48.90 &49.52 &44.21 &\textbf{47.73}/45.71/47.65 &\textbf{+1.17}/+0.51/+0.87  \\
 \textit{a+c+d} &49.42 &49.13  &49.68 &44.66 &47.72/\textbf{48.21}/47.94 &+1.16/\textbf{+1.24}/+0.96 \\
 \hline
  \textit{a$\times$3} &48.75  &48.09 &49.19 &43.18 &47.07/\textbf{47.28}/47.20 &+0.51/\textbf{+0.72}/+0.64  \\
  \textit{d$\times$3} &48.47  &49.02 &49.42 &44.09 &47.51/47.46/\textbf{47.57} &+0.73/+0.68/\textbf{+0.79}  \\
 \hline
   \textit{a+b+c+d} &49.52 &48.94 &49.61 &44.70 &47.95/46.10/\textbf{48.3}/47.95 &\textbf{+1.39}/+0.90/+1.33/+1.17   \\
   \hline
\end{tabular}
\end{center}
\caption{BLEU score for the representative models in multi-agent training on NIST Chinese-English translation. \textit{X$\times$Y} stands for $Y$ agents with the indentical model \textit{X} but from different initializing seeds.  }
\label{tbl:nist}
\end{table*}

\section{Experiments}
In this paper, we evaluate our model on four translation tasks: NIST Chinese-English Translation Task, IWSLT 2014 German-English Translation Task, WMT 2014 English-German Translation Task and large-scale Chinese-English Translation Task. 
\subsection{Data Preprocessing}
To compare with previous studies, we conduct byte-pair encoding \cite{DBLP:journals/corr/SennrichHB15} for Chinese, English and German sentences, setting the vocabulary size to 20K and 18K for Chinese-English, a joint 20K vocabulary for German-English, and a joint 32K vocabulary for English-German, respectively.

For Chinese-English task,
the training data consists of about 1.5M sentence pairs extracted from LDC corpora \footnote{LDC2002E18, LDC2002L27, LDC2002T01,
LDC2003E07, LDC2003E14, LDC2004T07, LDC2005E83,
LDC2005T06, LDC2005T10, LDC2005T34, LDC2006E24,
LDC2006E26, LDC2006E34, LDC2006E86, LDC2006E92,
LDC2006E93, LDC2004T08(HK News, HK Hansards )}. We choose the NIST 2006 (MT06) dataset for validation, and NIST 2002-2004 (MT02-04), as well as NIST 2008 (MT08) datasets for testing. For large-scale Chinese-English task, the training data consists of about 40M sentence pairs extracted from web data.

\subsection{Agent Variants}
To increase the diversity of agents, we implement the following variants of the Transformer based system: \textit{L2R}, the officially released open source toolkit for running Transformer model, \textit{R2L}, the standard Transformer decodes in reversed direction, \textit{Enc}, the standard Transformer with 30 layers in the encoder, and \textit{Rel}, the reimplementation of self-attention with relative position \cite{shaw2018self}.

\subsection{Training Details}
We implement our models using PaddlePaddle \footnote{\url{https://github.com/paddlepaddle/paddle}}, an end-to-end open source deep learning platform developed by Baidu. It provides a complete suite of deep learning libraries, tools and service platforms to make the research and development of
deep learning simple and reliable. 

We use the hyperparameters of the \textit{base} version for the standard Transformer model of NIST Chinese-English and IWSLT German-English translation tasks, except the smaller token size ($batch\_size=320$). For WMT English-German and large-scale Chinese-English translation tasks, we use the hyperparameters of the \textit{big} version. 
As described in the previous section, we use a hyperparameter $\lambda_i$ for each agent to balance the preference between learning from the observed data and the ensemble model. According to the performance of each agent after pre-training, we set this value as follows:
\begin{equation}
\lambda_i = 0.5 + \max (-0.5, \min(0.5, \frac{B_i - B_{avg}}{10}))
\label{eqn:lambda}
\end{equation}
where $B_i$ is the BLEU score obtained by the pre-training of the $i-th$ agent, and the $B_{avg}$ is the average BLEU score of all agents.

The above formula suggests the agent learns more from the ensemble model as its performance is worse than the majority vote, rather than focusing on exploring by its own prediction.
 
We train our model with parallelization at data batch level. %we penalize the number of steps by dividing the number of GPU we use. 
For NIST Chinese-English task, it takes about 1.5 days to train models on 8 NVIDIA P40 GPUs, 5 days for WMT English-German task, 8 hours for IWSLT German-English task and 7 days for large-scale Chinese-English task. The detailed training process is as follows, first we train each agent until BLEU doesn't improve any more and then execute  Algorithm \ref{algo1} for one-to-many learning which takes 30K-40K steps to converge. %In order to avoiding overfitting, dropout is also applied on the attention layer and fc layer and we set the dropout rate to 0.1. In practice, we use single agent for inference.
 
\subsection{Chinese-English Results}
\textbf{Study on different numbers of agents.} We first assess the impact of diverse agents on translation quality. From Table \ref{tbl:nist}, we see that the model's performance consistently improves as the number of agents increases, and we observe that 1) The four baseline systems with different implementations present diverse translation quality, in particular \textit{Rel} with refined position encoding achieves the best performance. 
%2) Even co-training with a worse agent (\textit{R2L}), the performance of each agent can be further improved ( \textit{L2R} (+0.33) in \textit{a+b};  \textit{L2R} (+1.07) and \textit{Enc} (+1.12) in \textit{a+b+c}; \textit{L2R} (+0.90) and \textit{Rel} (+0.75) in \textit{a+b+d}).
 2) The improvement of each agent after multi-agent learning is dependent on the performance of the co-trained agent (+0.33, +0.8, +0.63 improvements obtained by \textit{L2R} when training with \textit{R2L}, \textit{Enc} and \textit{Rel}). 3) Better results can be obtained by increasing the number of training agents (e.g., 47.3 $\rightarrow$ 47.73 $\rightarrow$ 47.95 for L2R).

From the overall results,  increasing the number of agents in multi-agent learning significantly improves the performance of each agent (at most +1.39, +0.9, +1.33, +1.17 in \textit{a+b+c+d}), which suggests that each agent learns advantages from the other agents, and more agents might lead to further improvement. 
More importantly, our improvements are obtained from the advanced training strategy without any modification in decoding stage, which indicates its practicability in deployment.  
\begin{table}[t]
\begin{center}
\begin{tabular}{l|c|c|c|c|c}
 Task& \textit{L2R} &\textit{Rel} & KD-4 & Dual-5 & Rel-4  \\
 \hline
 De-En &33.63  &34.91&35.53  &34.70& \textbf{36.27}  \\
   \hline
\end{tabular}
\end{center}
\caption{BLEU score on IWSLT 2014 German-English translation. KD-4 stands for ensemble knowledge distillation with four agents. Dual-5 is the SOTA model from the work of \citet{dual2019}. And Rel-4 is our best model (\textit{Rel}) training with four diverse agents.  }
\label{tbl:ge}
\end{table}

\noindent \textbf{Study on uniform agents.} To measure the importance of diversity in multi-agents, we conduct multiple experiments by initializing the similar model with different initialization seeds to generate multiple agents. As reported in Table \ref{tbl:nist}, although the single \textit{Rel} model achieves the best performance compared to the other three single agents, the \textit{d $\times$ 2} (47.44) and \textit{d $\times$ 3} (47.57) perform worse than the best counterparts \textit{c+d} (47.79) and \textit{a+c+d} (47.94). Moreover, when training with more than two agents, the performance of \textit{d $\times$ 3} (47.57) is even lower than arbitrary diverse counterparts (\textit{a+b+c} (47.74); \textit{a+b+d} (47.73)). 

These results suggest that multi-agent learning is better conducted with diverse agents instead of the uniform agents with excellent single performance. On the other hand, in our multi-agent learning, training with more agents brings consistent improvements even when deployed by identical models from different initialization seeds.

\subsection{German-English and English-German Results}
We work with the IWSLT German-English translation to compare our model to the SOTA model. From Table \ref{tbl:ge}, we observe that both KD-4 and Rel-4 models achieve the best result on this dataset (35.53 and 36.27), which manifests the effectiveness of using multiple agents. 

Even when trained with four agents, our model outperforms the dual model trained with five agents (34.70), and reports a SOTA score on this translation task. Moreover, we argue that fine-tuning with \textit{L2R} to obtain a better performance could further improve the performance of Rel-4, as there exists a large gap between \textit{L2R} and \textit{Rel}, which affects the learning efficiency.

\begin{table}[t]
\begin{center}
\begin{tabular}{l|c}
 Models & En-De  \\
 \hline
 ConvS2S \citet{gehring2017convolutional} & 25.2 \\
 Transformer \citet{vaswani2017attention} & 28.4\\
 \textit{Rel} \citet{shaw2018self} & 29.2\\
 DynamicConv \citet{wu2019pay} & 29.7 \\
 \hline
 Back-translation \citet{edunov2018back}& 35.00 \\
 Dual-3 \citet{dual2019} & 29.92 \\
  Dual-3 + Mono Data& 30.67 \\
 \hline
 \textit{L2R} & 28.37 \\
 \textit{Rel} & 29.16\\
 
 Rel-4& 29.67  \\
   \hline
\end{tabular}
\end{center}
\caption{BLEU score on newstest2014 for WMT English-German translation.   }
\label{tbl:eg}
\end{table}

We further investigate the performance of our model on the WMT English-German translation task, which achieves a competitive result of 29.67 BLEU score on newstest2014. From Table \ref{tbl:eg}, we can see that another related work, multi-agent dual learning \cite{dual2019} achieves promising results. This confirms that training with more agents leads to  better translation quality, and reveals the relevance of using multiple agents.

Although \citet{edunov2018back} reports a BLEU score of 35.0 on this dataset, they leverage refined training corpus and a large number of monolingual data. We argue that our model can bring further improvement using their back-translation technique. 
Moreover, the goal of this paper is introducing a general learning framework for multi-agent learning rather than exhaustively fine-tuning to report a SOTA results. We argue that the performance of our model can be further improved using an advanced single agent, such as DynamicConv \cite{wu2019pay} and Dual \cite{dual2019}.

\begin{table}[t]
\begin{center}
\begin{tabular}{l|c}
 Models &   AVERAGE\\
 \hline
 \textit{L2R(baseline)} & 33.74 \\
 \textit{L2R+R2L+Enc+Rel} & \textbf{34.60}\\
   \hline
\end{tabular}
\end{center}
\caption{BLEU score for the L2R model in multi-agent training on large-scale corpus of Chinese-English translation. Because we only use L2R model in Baidu Translate, so part of results are presented. }
\label{tbl:liju}
\end{table}

\subsection{Large-Scale Chinese-English Results}
In order to prove that multi-agent learning can yield improvement in large-scale corpus, we conducted experiments on 40M Chinese-English sentence pairs with the same data processing method as NIST. We built test sets of more than 10K sentences which cover various domains such as news, spoken language and query logs. From Table \ref{tbl:liju}, we can see that multi-agent learning can increase the average baseline BLEU score of several test sets by 0.86. This technique has already been applied to Baidu Translate.

\subsection{Contrastive Evaluation}
%Contrastive evaluation \cite{sennrich2017grammatical} tests the sensitivity of NMT models to specify translation errors. The contrastive examples are designed to capture specific translation errors rather than to evaluate the global quality of NMT models. Although they do not replace metrics such as BLEU. They give further insights into the performance of models on specific linguistic phenomena. %
In this paper, we evaluate our model on the two types of contrastive evaluation datasets: 
\textit{\textbf{Lingeval97}} \footnote{\url{https://github.com/rsennrich/lingeval97}} and \textbf{\textit{ContraWSD}} \footnote{\url{https://github.com/a-rios/ContraWSD}}. The former is utilized to measure the performance of different models on dealing with subject-verb agreement in English-German translation, while the latter evaluates the performance on word sense disambiguation in German-English translation. We suggest the readers refer to the work \cite{sennrich2017grammatical,D18-1458} for detailed descriptions of the two datasets.

From Table \ref{tbl:contrastive}, we can see that the \textit{L2R} training with multiple agents improves the accuracy both on SVA and WSD, and we observe that: 1) The \textit{L2R} improves its ability of SVA with the help of the other agents significantly. 2) The \textit{Enc} presents better advantage in resolving WSD than in SVA, while \textit{Rel} has the opposite preference. 3) Although the \textit{R2L} presents ordinary BLEU score, it still does help the \textit{L2R} resolve the SVA.
\begin{table}[t]
\begin{center}
\begin{tabular}{l|c|c}
 Models & SVA (\%)  & WSD (\%)\\
 \hline
 \textit{L2R} &89.06 &78.05\\
  \textit{R2L} &88.92 &77.97\\
   \textit{Enc} &89.13 &\textbf{79.07}\\
    \textit{Rel} &\textbf{89.21} &78.75\\
 \hline
  \textit{L2R+R2L}  &89.28 &77.94\\
   \textit{L2R+Enc}  &\textbf{89.42} &78.29\\
     \textit{L2R+Rel}  &89.39 &\textbf{78.47}\\
 \hline
 \textit{L2R+R2L+Enc}  &89.26 &78.36\\
  \textit{L2R+R2L+Rel}  &\textbf{89.58} &78.28\\
  \textit{L2R+Enc+Rel}  &89.39 &\textbf{78.84} \\
\hline
 \textit{L2R+R2L+Enc+Rel}  &89.47 &78.62\\
\end{tabular}
\end{center}
\caption{Accuracy of subject-verb agreement (SVA) and word sense disambiguation (WSD) for different models. We report the performance of \textit{L2R} in different models for comparison of using different agents.  }
\label{tbl:contrastive}
\end{table}
\section{Related Work}
The most related work is recently proposed by \citet{dual2019}, who introduced a multi-agent algorithm for dual learning. However, the major differences are 1) the training in their work is conducted on the two agents while fixing the parameters of other agents. 2) they use identical agent with different initialization seeds. 3) our model is simple yet effective. Actually, it is easy to incorporate additional agent trained with their dual learning strategy in our model, to further improve the performance. More importantly, both of our work and their work indicate that using more agents can improve the translation quality significantly.
%their work is actually training on two agents with different initialization seeds, and constrains in dual model, it leverages the duality among agents to learn a better model. 

%Actually, we can also incorporate an additional agent trained with dual learning in our model, to further improve the performance of our model.

Another related work is ensemble knowledge distillation \cite{fukuda2017efficient,freitag2017ensemble,P18-1129, zhu2018knowledge}, in which the ensemble model of all agents is leveraged as the \textit{Teacher} network, to distill the knowledge for the corresponding \textit{Student} network. However, as described in the previous section, the knowledge distillation is one particular case of our model, as the performance of the \textit{Teacher} in their model is fixed, and cannot be further improved by the learning process.

Our work is also motivated by the work of training with two agents, including dual learning \cite{he2016dual,xia2017dual,xia2018model},  and bidirectional decoding \cite{liu2016agreement,zhang2018asynchronous,zhang2018regularizing,zhang2019synchronous}. Our method can be viewed as a general learning framework to train multiple agents, which explores the relationship among all agents to enhance the performance of each agent efficiently.

\section{Conclusions and Future Work}
In this paper, we propose a universal yet effective learning method for training multiple agents. 
In particular, each agent learns advantages from the ensemble model when necessary. The knowledge distillation from the ensemble model to the agent stems the better translation generated by the ensemble model, which enables each agent to learn high-quality knowledge while retaining its own exploration. 

Extensive experimental results prove that our model brings an absolute improvement over the baseline system, reporting SOTA results on IWSLT 2014 German-English and competitive results on WMT 2014 English-German translation tasks.

In the future, we will focus on training with more agents by translating apart of one sentence considering its advantage for each agent, rather than translating the whole sentence. 

\section{Acknowledgements}
We would like to thank Ying Chen, Qinfei Li and the anonymous reviewers for their insightful comments.

\bibliography{emnlp-ijcnlp-2019}
\bibliographystyle{acl_natbib}

\end{CJK}
\end{document}